# Development of Control Framework for Spine Surgery Robot Using EtherCAT

Veysi ADIN
Center for Healthcare Robotics
Korea Institute of Science and Technology
Division of Nano and Information Technology
University of Science and Technology
Daejeon, South Korea
veysi.adin@kist.re.kr

Chunwoo Kim
Center for Healthcare Robotics
Korea Institute of Science and Technology
Division of Nano and Information Technology
University of Science and Technology
Daejeon, South Korea
cwkim@kist.re.kr

*Abstract*—As the more sensors and actuators are used in the robotic systems to provide more features, complexity of the system is increasing. When it comes to medical robotics, it becomes harder to ensure safety and determinism in the system. To deal with increasing complexity and ensure precise periodicity and execution timing for a medical robot, in this paper we report development of EtherCAT master as a part of software framework for spine surgery robot. We implemented multi-axis controller using open-source EtherCAT master running in real-time preemptive Linux. We evaluated the real-time performance of the system in terms of periodicity, jitter and execution time in our first prototype of spine surgery robot.

## I. Introduction

Development of robot requires integration of various mechanical electrical and software components. For medical robots, system integration process needs to consider general standards for medical devices and software such as IEC-60601 [1] (safety and essential performance of medical electrical equipment standard), IEC-62304 [2] (medical device software life cycle standard) and IEC80601-2-77 [3] (Medical electrical equipment — Part 2-77: Particular requirements for the basic safety and essential performance of robotically assisted surgical equipment). To comply with these standards, in terms of physical integration, high speed, reliable, deterministic communication protocol is required.

Physical integration is process of establishing a physical connection between the components and providing a communication method through this connection. Various fieldbus and protocols such as Controller Area Network (CAN), RS232, RS485, Universal Serial Bus (USB), IEEE 1394 (Firewire), and EtherCAT, are used for physical integration of medical robot [4], [5]. As complexity and dynamic requirement of the robot system increases, fieldbus with high bandwidth for reliable and fast communication, is necessary. EtherCAT is a fieldbus providing fast (up to 10Gbps) and flexible protocol that uses standard Ethernet card for IO. Though relatively new, its user base is increasing and many commercial actuators and IO boards are now available with EtherCAT interfaces.

In this paper, as a part of a larger effort for developing a software framework that can be used for various medical robot and device development, we present implementation of controller of a spine surgery robot system. The controller uses EtherCAT as a fieldbus for physical integration with master running on real-time preemptive Linux. Implementation details and realtime performance evaluation by jitter measurement is presented in subsequent sections.

## II. Implementation Details

### A. EtherCAT master implementation

EtherCAT is an Ethernet based fieldbus standard focusing on real-time requirements for automation, and it is designed to achieve very low cycle times typically less than 100 µs with low jitter less than 1 µs [6]. Many higher layer protocols such as CANopen over EtherCAT (CoE), File Access over EtherCAT (FoE) etc. [6] are available. Additionally, many robot hardware including I/O boards, motor drivers, sensors, and actuators with EtherCAT interface readily are available in the market. To control these component through EtherCAT, a computer running EtherCAT master is required to moderate communications. In this research, open-source implementation of EtherCAT master, IgH EtherLab [7], is used.

### B. Real-time Linux

For fast and reliable EtherCAT communication, EtherCAT master needs to run on real-time operating system. Linux itself is not real-time, however there are several methods to bring real-time features to Linux such as dual-kernel approach with Xenomai and RTAI, or single-kernel approach with RT_PREEMPT maintained by Linux Foundation.Even though dual kernel implementations has better real-time performance [8], considering implementation time, portability, scalability, adaptation to programming environment of each patch, and community support, RT_PREEMPT is better real-time solution for our application. Therefore, our software integration part consists of real-time patched (RT_PREEMPT) Linux distribution Xubuntu 20.04.

This work was supported by the Technology Innovation Program (20009071) funded By the Ministry of Trade, Industry and Energy (MOTIE, Korea)

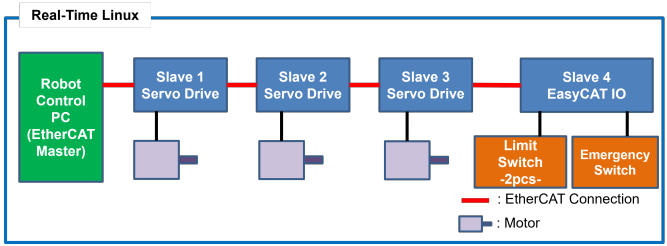

Figure 1. EtherCAT Connection Diagram

Table I
SOFTWARE & HARDWARE IMPLEMENTATION DETAILS

| Master (Advantech MIO-5272 Embedded Single Board Computer) | | | |
|---|---|---|---|
| CPU | RAM | Network Card | |
| Intel Core i7-6600U CPU @ 3.2GHz | 8GB | Intel I219 Network Card (driver: generic) | |
| OS | Kernel + RT_Patch | Task Priority | Etherlab Version |
| Xubuntu 20.04 | 5.10.73-rt54 | 98 | v1.5.2 |
| Slaves | | | |
| Servo Drives | | Custom Slave | |
| Maxon Epos4 50/5 Compact -3 pcs.- | | EasyCAT Custom Slave - 1 pcs.- | |
| PDO Mappings Servo Drives | | PDO Mappings Custom Slave | |
| 28 bytes each, TxPDO 14 bytes RxPDO 14 bytes | | 8 bytes, TxPDO 4 bytes RxPDO 4 bytes | |
| Control Frequency | | 1 KHz | |

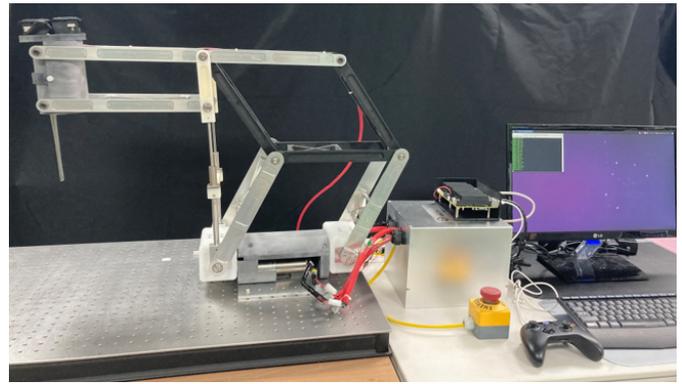

Figure 2. Experimental Setup

Table II
REAL-TIME PERFORMANCE EVALUATION EXPERIMENT RESULTS

| | Avg. ± St.D | |
|---|---|---|
| $T_{period}$ (μs) | Avg. ± St.D | 1000.017±7.035 |
| | Min / Max | 870.388 / 1148.032 |
| $T_{jitter}$ (μs) | Avg. ± St.D | 2.125±6.706 |
| | Min / Max | 0 / 148.032 |
| $T_{exec}$ (μs) | Avg. ± St.D | 11.479±7.039 |
| | Min / Max | 6.194 / 231.767 |

## III. EXPERIMENTAL RESULTS

Multi-axis controller using open-source EtherCAT master running in real-time preemptive Linux was implemented and was used to control the prototype of spine surgery robot shown in Fig. 2. Three motors and two limit switch and emergency button of the robot is connected through EtherCAT as shown in Fig. 1. Implementation details in terms of hardware and software shown in Table I.

Performance of the implemented controller was measured by measuring jitter, periodicity and execution time of our real-time loop. Controller input acquired from Xbox controller via USB communication in second thread is passed into real-time communication thread running at 1 kHz, and communication thread sends desired velocity to drivers in synchronous cylic velocity mode. Controller was left running for three hours and timing samples were acquired every 10ms which created 1.08 million timing samples. Acquired timing results shown in Table II. According to the table, maximum jitter of our implementation is 148.032 μs and maximum execution time is 231.767 μs. Timing measurements showed that, our EtherCAT based integration method is suitable for real-time applications.

## IV. CONCLUSION

In this paper, we reported development of control framework for spine surgery robot using EtherCAT protocol for physical integration. We measured real-time performance of our Ether-CAT master implementation and, showed that it is suitable for real-time applications.

As a future work, we will be implementing our framework by using robotic middlewares, with respect to component based software engineering.